\pdfoutput=1

\documentclass[11pt]{article}

\usepackage[final]{acl}

\usepackage{times}
\usepackage{latexsym}

\usepackage[T1]{fontenc}

\usepackage[utf8]{inputenc}

\usepackage{microtype}

\usepackage{inconsolata}

\usepackage{graphicx}
\usepackage{xcolor}
\usepackage{amsmath}
\usepackage[ruled]{algorithm2e}
\usepackage{booktabs}
\usepackage{colortbl}
\usepackage{tcolorbox}
\usepackage{enumitem}
\usepackage{roboto}  
\usepackage{lipsum}  
\usepackage{tabularx}  
\usepackage{float}

\title{Advancing Healthcare Automation: Multi-Agent System for Medical Necessity Justification}

\author{Himanshu Pandey \\
  RISA Labs \\
  \texttt{himanshu@risalabs.ai} \\\And
  Akhil Amod \\
  RISA Labs \\
  \texttt{akhil@risalabs.ai} \\\And
  Shivang \\
  RISA Labs \\
  \texttt{shivang@risalabs.ai}
  }

\begin{document}
\maketitle
\begin{abstract}
Prior Authorization delivers safe, appropriate, and cost-effective care that is medically justified with evidence-based guidelines. However, the process often requires labor-intensive manual comparisons between patient medical records and clinical guidelines, that is both repetitive and time-consuming. Recent developments in Large Language Models (LLMs) have shown potential in addressing complex medical NLP tasks with minimal supervision. This paper explores the application of Multi-Agent System (MAS) that utilize specialized LLM agents to automate Prior Authorization task by breaking them down into simpler and manageable sub-tasks. Our study systematically investigates the effects of various prompting strategies on these agents and benchmarks the performance of different LLMs. We demonstrate that GPT-4 achieves an accuracy of 86.2\% in predicting checklist item-level judgments with evidence, and 95.6\% in determining overall checklist judgment. Additionally, we explore how these agents can contribute to explainability of steps taken in the process, thereby enhancing trust and transparency in the system.
\end{abstract}

\section{Introduction}

In US healthcare, management of administrative workflows represents a pivotal yet formidable challenge. Physicians, nurses, and administrative personnel frequently allocate a substantial portion of their working hours to these procedural tasks, distracting from their primary focus on patient care. One such workflow, Prior authorization (PA) is a healthcare management process used by insurance entities to determine whether a proposed treatment or service is covered under a patient’s plan before it is approved to be carried out. This process applies to various treatments and services, including medications, imaging, and procedures \cite{Madhusoodanan2023}. Evaluating a PA application involves assessing medical necessity of patient-specific health records against prevailing coverage guidelines. A major part of these coverage guidelines are clinical guidelines which are systematically developed statements designed to help practitioners make decisions about appropriate health care for specific clinical circumstances. Insurance companies review these clinical guidelines to to justify medical necessity of a procedure or treatment \cite{chambers}.

While Prior Authorization ensures safe, appropriate, cost-effective and evidence based care to all members \cite{Jones2021}, it is a major source of physician and staff burnout as well as job dissatisfaction.There are several ongoing efforts to improve the prior authorization process. High-profile innovations include (1) “gold carding” providers, exempting those who have very high historical approval rates; and (2) automating the process through e-prior auth (e-PA) \cite{Lenert2023}. e-PA proposes a set of transactions conveying the rules for approval in a standardized query representation in CQL. While such rule based methods are adequate for simple authorization decisions, complex cases with temporal data, evidence of responses and trends in clinical data items can be difficult to represent in CQL's rule based format \cite{Lenert2023}. Also, a nationwide survey \cite{Salzbrenner2022} identified that the use of e-PA was not associated with less provider time or fewer challenges in preparing and submitting PA requests. However, the use of e-PA reported a shorter PA decision time. Additionally, there is an understanding that AI can potentially improve the current state of PA filing \cite{Lenert2023}.

The introduction of Large Language Models (LLMs) \cite{openai2024gpt4, touvron2023llama} has catalyzed a transformative shift in the capabilities of artificial intelligence, enabling the resolution of complex challenges previously inaccessible to conventional AI methods. LLMs excel in interpreting and synthesizing large volumes of unstructured data, enhancing tasks such as natural language understanding \cite{yang2024harnessing}, sentiment analysis, and automated content creation \cite{zhou2024survey}. Building on this foundation, \textit{Multi-Agent Systems}, which employs a collective of AI-powered agents, represents an even further advancement \cite{guo2024large}. This approach decomposes a singular complex task into multiple, manageable sub-tasks and distributes them across multiple agents, each specialized through training for a sub-task. Following this methodology essentially infuses a microservice architecture into the traditional monolithic AI framework, enabling more modular, scalable, and robust AI systems. By integrating the depth and adaptability of LLMs with the collaborative and dynamic nature of Multi-Agent Systems, AI systems can achieve unprecedented levels of performance and versatility across various complex problems \cite{guo2024large, he2024llm}.

In this paper, we investigate the application of multi-agent systems for determining medical necessity for a medical procedure. Our contributions are as follows:

\begin{itemize}
  \item We propose a novel challenge of establishing medical necessity for prior authorizations (PAs) by reasoning on clinical guidelines against patient medical records.
  \item We decompose the problem statement of PA filing into intermediate sub-tasks, which can then be effectively solved by LLM Agents.
  \item We demonstrate through extensive experimentations the effect of LLM choice and prompting strategies. Specifically, GPT-4 achieves an accuracy of 86.2\% in predicting checklist item-level judgments and 95.6\% in determining overall checklist judgment.
\end{itemize}
\section{Related Work}

Large Language Models (LLMs) have completely changed the landscape of Natural Language Processing (NLP) in the recent years. LLMs have shown \textit{emergent abilities} \cite{wei2022emergent} in settings like few-shot prompting \cite{brown2020language} and augmented prompting strategies. Augmented prompting like Chain of Thought (CoT) \cite{NEURIPS2022_9d560961} and Automatic Chain of Thought \cite{zhang2022automatic} prompting enables LLMs to solve reasoning tasks using step by step approach. Additionally, instruction fine-tuning with human feedback has made LLMs able to respond to instructions describing unseen tasks \cite{ouyang2022training}. Other advancements include techniques like self-consistency \cite{wang2023selfconsistency} which helps LLMs solve complex tasks using multiple different ways of thinking and prompt gradient descent \cite{pryzant2023automatic} which edits prompt in the opposite semantic direction of the gradient to boost prompt's performance. Building on this, more dynamic and complex tasks can be tackled by LLM powered Multi Agent Systems (LLM-MAS). These LLM-MAS have collaborative autonomous agents equipped with unique strategies and behaviour \cite{guo2024large}. This agentic behaviour is based on the idea that LLMs can improve in game-play scenario by using previous experiences and feedback \cite{fu2023improving, madaan2023selfrefine}.

LLMs have the potential to disrupt medicine. Models like Med-PaLM \cite{singhal2022large} outperformed state of the art on all MultiMedBench tasks \cite{doi:10.1056/AIoa2300138}. GPT-4 has consistently outperformed task-specific fine-tuned models and is comparable to human experts on QA datasets \cite{zhou2024survey}. GPT-4 scored 86.65\% in United States Medical Licensing Examination (USMLE) where passing percentage was 60\% \cite{nori2023capabilities}. It also demonstrates  GPT-4’s capacity for reasoning about concepts tested in USMLE challenge problems, including explanation, counterfactual reasoning, differential diagnosis, and testing strategies. Some recent researches have started to explore the impact of LLMs in discharge summary generation \cite{ellershaw2024automated, williams2024evaluating}, care planning \cite{nashwan2023enhancing, jung2024enhancing}, Electronic Health Records (EHRs) \cite{cui2024multimodal, ahsan2023retrieving}. Text-to-SQL parsing has attracted significant interest \cite{li2024can}. Building on this idea, numerous research efforts, such as EHRSQL \cite{EHRSQL}, are focused on extracting data from EHRs. Additionally, there are ongoing efforts to develop solutions for EHR-based question-answering tasks \cite{shi2024ehragent}.

\begin{figure*}
    \centering
    \includegraphics[width=\textwidth]{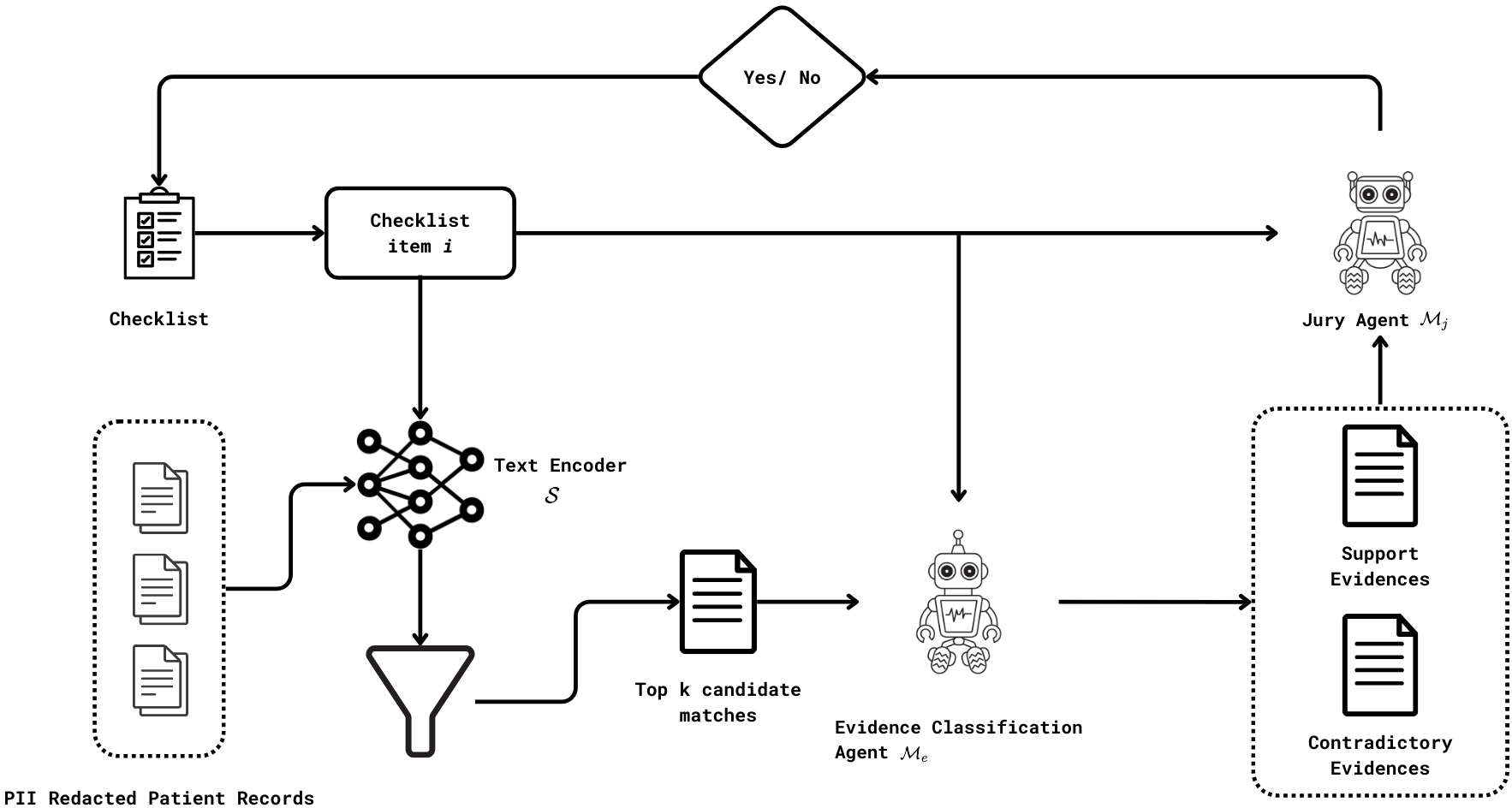}
    \caption{Leaf-Level Judgement Prediction where the first agent classifies the documents into supporting and contradictory sets and then the jury agent determines if the checklist item is satisfied.}
    \label{fig:agent1}
\end{figure*}

However, the domain of PA filing is largely untouched by LLMs mainly because of lack of publicly available data despite the understanding that AI can potentially improve its current state \cite{Lenert2023}. While some efforts have been made to automate PA filing, for example \cite{Diane2023} where ChatGPT is utilised to generate PA letters for Orthopedic Surgery Practice, but the process lacks the important step of establishing medical necessity using AI. Another study aims to determine PA Approval for Lumbar Stenosis Surgery with Machine Learning \cite{DeBarros2023} but it uses surgery specific symptoms as input variables which would be difficult to generalize.
\section{Problem Statement}

As mentioned above, the evaluation of medical necessity is conducted through a meticulous comparison between patient medical records and established clinical guidelines. These medical records are systematically structured in a json-like format, usually in FHIR \footnote{https://www.hl7.org/fhir/}, within Electronic Health Records (EHRs) systems. Each object (resource) can be of type Patient (Patient Demographics), Observation (Laboratory Results), Procedure (Treatment History), Medication Request, Diagnostic Report etc. We define a set of EHR documents (resources) as $\mathcal{D} = \{d\}_{i=1}^{N_D}$ of size $N_D$

\begin{figure*}
    \centering
    \includegraphics[width=\textwidth]{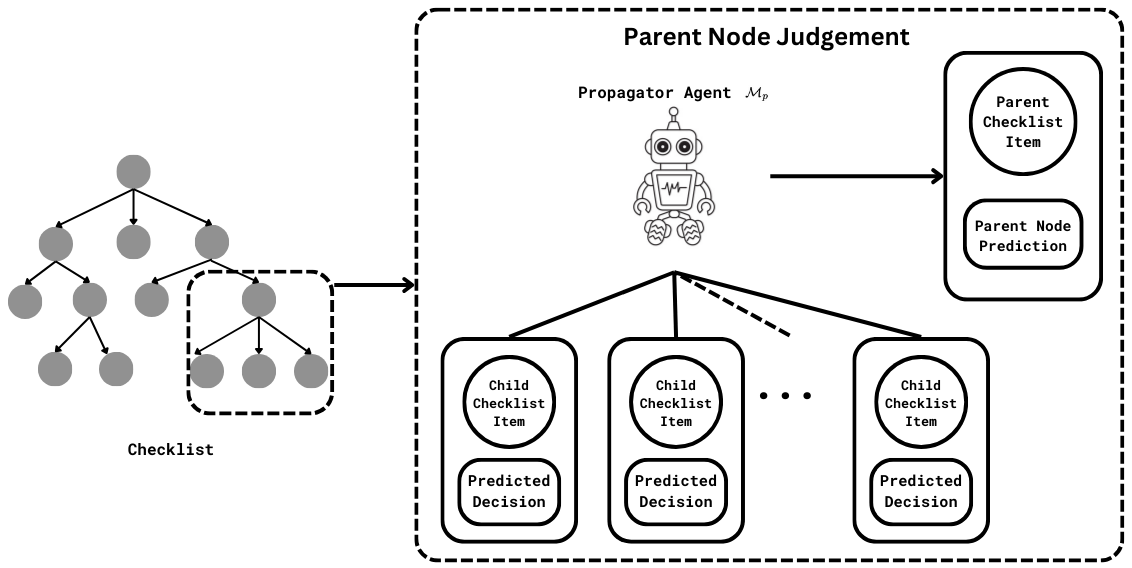}
    \caption{Bottom-Up Judgement Propagation where the agent uses the logical operators contained in a checklist item to determine how the aggregation should take place.}
    \label{fig:agent2}
\end{figure*}

Further, clinical guidelines are formatted in a hierarchical, tree-like structure (referred as \textit{checklist} in this paper), where each guideline statement (\textit{parent node}) can encompass an arbitrary number of subordinate child statements (\textit{sub-checklist or leaf node}) nested within it as shown in Figure \ref{fig:agent2} and \ref{fig:footwear}. Thus, we define a coverage guideline or checklist as $\mathcal{C} = \{c\}_{j=1}^{N_C}$, where $c$ is a checklist item.

Eventually, the task is to automatically determine the medical necessity $Y \in \{-1,0,1\}$ where -1 means the medical necessity is not justified, 1 means it is justified and 0 means there is a lack of sufficient evidence to justify the medical necessity criteria. 

Recognizing the importance of transparency in the task, we also aim to provide evidence $\mathcal{E}_c = \{e_{c_k}\}_{k=1}^{N_c}$. These evidences can be used downstream to cross-reference medical documents used to establish medical necessity for the procedure.

We aim to construct a machine learning model $\mathcal{M}$ such that:
\begin{equation}
    \mathcal{M}(\mathcal{D}, \mathcal{C}) = \{Y, \{\mathcal{E}_{c}\}\}\  \forall c \in \mathcal{C}
\end{equation}
\section{Methodology}

Recently Large Language Models have shown great performance improvements by breaking down complex tasks into simpler sub-problems \cite{khot2022hey}. Motivated by this observation, we propose a two step solution for our problem statement. First we determine the judgement of each of the leaf node checklist item. Subsequently, we propagate the solution for parent nodes bottom-up based on its child nodes' judgements.

\subsection{Leaf-Node Judgement Prediction}
 Considering the immense volume of documents in Electronic Health Records (EHRs), we propose a Retrieval-Augmented Setup \cite{gao2024retrievalaugmented}. This approach first filters the document pool to identify a set of likely evidences (\textit{top-k evidences}). A \textit{Classification Agent} is then utilized to select the relevant evidences for the specific checklist item, enabling precise and efficient data extraction.

\paragraph{Top-k Evidence Selection:} Given the EHR data \(\mathcal{D}\), we first decompose it into its constituent resources (documents) where each document is an individual entry (individual lab-report data, procedure etc.) . In order to filter-down documents that are redundant towards the judgment, we first obtain top $k$ candidate matches for the checklist item $c$ from $D$. To achieve this, we propose to use a text encoder $\mathcal{S}$ to derive semantic representations for each checklist item $c$ and for each document $d$ in the EHR data. This method allows us to map both the checklist items and the documents into a shared semantic space, facilitating more effective matching based on relevance. Subsequently, we employ a semantic similarity metric to calculate the similarity score between each document $d$ and the checklist item $c$. Based on the similarity metric, we obtain top-$k$ closest matched documents with the checklist item $c$. Note that due to cost involved in using LLMs for this task, we keep\footnote{In our experiments section, we show how the performance of our approach varies with $k$} $k<50$. 

\begin{equation}
    \mathcal{S}(\{d_i\}|_{i=1}^{N_D}, c ) = \{d_{i'}\}|_{i'=1}^{k}\   \forall c \in \mathcal{C}
\end{equation}

\paragraph{Evidence Retrieval and Prediction:} Our proposed \textit{Evidence Classification Agent} $\mathcal{M}_e$, first looks at each document $d_{i}$ in top-$k$ evidences retrieved along with the checklist item $c$ and gives a verdict $v_i$, whether the document $d_{i}$ is a supporting evidence, a contradictory evidence or it does not affect the judgment $y_{c}$. Note that this agent is executed $k$ times since there are $k$ retrieved documents.
\begin{equation}
    \mathcal{M}_e(\{d_i, c\}|_{i=1}^{k}) = \{v_{i}\}|_{i=1}^{k} \   \forall c \in \mathcal{C}
\end{equation}

Then our \textit{Jury Agent} $\mathcal{M}_j$ picks up the complete set of evidences $d_{i}|_{i=1}^{k}$ along with their verdicts $v_{i}|_{i=1}^{k}$ and predicts the leaf-level checklist item judgment $y_{c}$ along with evidences $\mathcal{E}_{c} \subset s_{i}|_{i=1}^k$ that acted in favour of the judgement $y_{c}$. We run this leaf-node pipeline multiple times ($n=10$) and take vote of all predictions to determine the final judgement $y_c$. Confidence score $f_c$ is calculated as the percentage of times the majority answer is predicted by the agent. 

\begin{equation}
    \mathcal{M}_j(\{d_i, v_i\}|_{i=1}^{k},\ c) = \{y_{c}, f_c, \{\mathcal{E}_{c}\} \}  \   \forall c \in \mathcal{C}
\end{equation}

\subsection{Parent-Node Judgement Prediction}

The value of a parent node is contingent upon the values of its nested child nodes. Hence, we determine parent node's value by aggregating children nodes' values which are connected through logical operators (AND, OR, NOT).

\paragraph{Bottom-Up Judgement Propagation:} In order to obtain the decision over the complete checklist $\mathcal{C}$, we propose to use an iterative bottom-up approach. In another words,  we start from the leaf nodes and keep obtaining the judgment of their parent nodes. The iterations are terminated when we obtain the judgment and scores of the root node in the checklist $\mathcal{C}$. 

Mathematically, at every iteration $i$, we choose a set of leaf node checklist items $c_{j}|_{j=1}^{N_{par}}$ having a common parent checklist item $c_{par}$, having judgements $y_{j}|_{j=1}^{N_{par}}$ and confidence scores $f_{j}|_{j=1}^{N_{par}}$. We then calculate the judgement $y_{par}$ and confidence score $f_{par}$ of parent node as:
\begin{equation}
    \mathcal{M}_p(c_{par},\{c_j, y_j, f_j\}|_{j=1}^{N_{par}}) = \{y_{par}, f_{par}\}  
\end{equation}
where $ \mathcal{M}_p$ is our \textit{Propagator Agent}.

\section{Data Collection and Annotation}

Getting live EHR data for the purpose of this evaluation is difficult, costly and full of regulatory requirements. We therefore used de-identified discharge summaries from MIMIC-IV-Note \cite{Johnson2023MIMICIVNotes} as a proxy for this data. All discharge summaries therein have sections like chief complaint, history of present illness, past medical history, social history, physical and lab examinations, medications etc. which serves as the ideal data for this experiment. An average discharge summary has approximately 300 sentences divided into different categories. Joining this data with MIMIC-IV \cite{Johnson2023MIMICIV}, we can get the CPT/ICD-10 codes associated with each note. We also collected a set of publicly available clinical guidelines (from CMS etc.) pertaining to Cardiology and Oncology and cross referenced the CPT codes in these guidelines to our notes data, thus creating a dataset of (note, guideline) pairs.

An example checklist \footnote{Taken from CMS: https://www.cms.gov/medicare-coverage-database/search.aspx} is shown in Figure \ref{fig:footwear}. The checklist shows the clinical guideline for Therapeutic Footwear which consists of two items associated by \textbf{AND} operator. Item 2 in itself is a sub-checklist and will be true if any of the sub-checklist item is True as all of them are connected by \textbf{OR} operator.

\begin{figure}[ht]
\begin{tcolorbox}[
    arc=1mm, 
    colback=gray!10, 
    colframe=gray!80, 
    boxrule=0.5pt,
    fonttitle=\bfseries\scriptsize, 
    title=Example Checklist,
    fontupper=\scriptsize\fontfamily{pcr}\selectfont, 
    left=3mm, 
    right=3mm, 
    top=2mm, 
    bottom=2mm 
]
\raggedright
\textbf{Eligibility Checklist for Therapeutic Footwear}

\begin{enumerate}[leftmargin=*]
  \item The beneficiary has diabetes mellitus; and
  \item The certifying physician has documented in the beneficiary's medical record one or more of the following conditions:
    \begin{enumerate}
      \item Previous amputation of the other foot, or part of either foot;
      \item History of previous foot ulceration of either foot;
      \item History of pre-ulcerative calluses of either foot;
      \item Peripheral neuropathy with evidence of callus formation of either foot;
      \item Foot deformity of either foot;
      \item Poor circulation in either foot;
    \end{enumerate}
\end{enumerate}
\end{tcolorbox}
\captionof{figure}{An example checklist formatted as a decision tree}
\label{fig:footwear} 
\end{figure}

\subsection{Leaf Node Data Annotation}

\begin{figure}[h]
    \centering
    \includegraphics[width=\columnwidth]{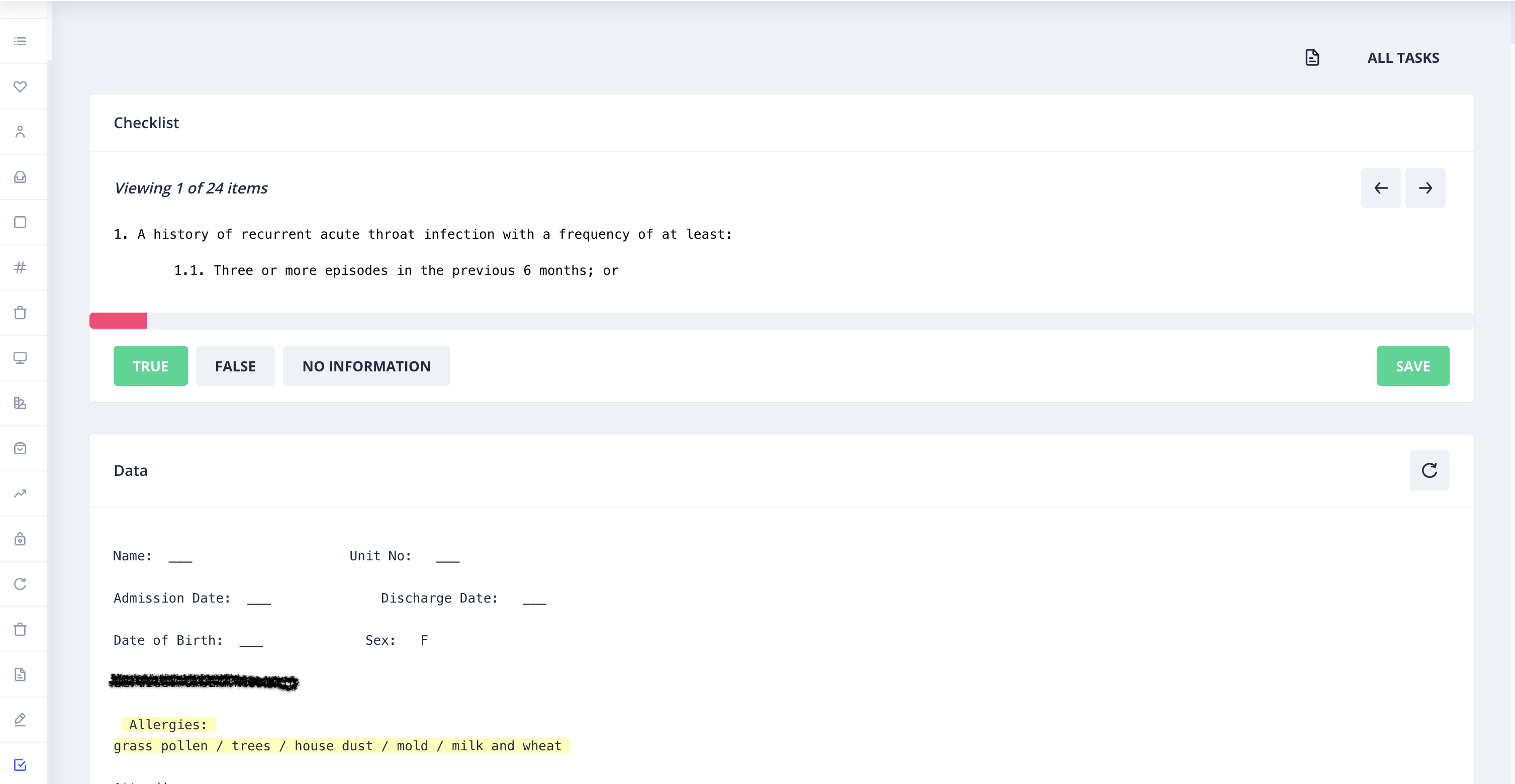}
    \caption{Annotation Dashboard where each annotator has to mark if the checklist item is True, False or No Information (can't be concluded) and mark evidences for their selection.}
    \label{fig:annotation}
\end{figure}

We hired 10 individuals with experience between 6-10 years in PA filing/reviewing both on payer and provider side. They were assigned the task of annotating leaf nodes of a checklist as either True, False or No Information. In case of True and False, the annotator has to also highlight statements in the data section as the evidences for that checklist item as shown in Figure \ref{fig:annotation}. Additionally, each (note, guideline) pair was annotated by 3 different annotators and the final verdict was determined by taking the majority vote of all annotators for that checklist item. Following this, we created a dataset of 281 annotated checklists having 4577 leaf checklist items. 

\subsection{Synthetic Data for Parent Judgement}

To test parent-node judgment propagation, we created synthetic data. This was needed because the logical operations required were not within the expertise of our medical domain annotators. To create synthetic data, we first extracted out all sub-checklists from the unique set of guidelines we had, and then manually labelled each sub-checklist with the operator (AND, OR and NOT) used for the aggregation of result for that sub-checklist. Then we randomly assigned each leaf node in all sub-checklist their judgements and confidence score and calculated the judgement and confidence score of the parent node programmatically. With different permutations of True, False and No Information used for each sub-checklist, we created a dataset of 4500 sub-checklists used for the evaluation of parent node judgement propagation. This method of synthetic assignment is advantageous as it introduces a range of less likely or extraordinary judgment combinations, thereby challenging our Propagator Agent to maximize its robustness.
\section{Experiments and Results}

Our experiments were categorized into two distinct segments: assessment of leaf-node judgment and evaluation of parent-node judgment. To facilitate this, we established two separate test environments. Each test-bed was equipped to integrate various Large Language Models (LLMs) to ascertain the optimal model for our needs.

\subsection{Leaf-node Judgement}

Leaf-node judgment encompassed three sequential tasks. We start by splitting the entire document into sentences. Note that, with MIMIC data it is an easy way to chunk EHR data, but in real case scenario the chunking would happen at FHIR resource \footnote{Refer: https://www.hl7.org/fhir/} level i.e. each Observation, Encounter, Lab Data etc. will act as the smallest chunk that goes into the pipeline. These chunks (or sentences here for simplicity) is first passed through the an encoder module which sorts the sentences according to the cosine similarity. The first 20 sentences are chosen for the experimentation. This simplifies the task of Classification Agent and also saves on LLM cost. The classification agent then segregates these filtered sentences in group of supporting and contradictory evidences which helps predicting the final judgement $y_c$ by the Jury Agent.

Note that the evidences given by the model for each checklist item is not generated but classified. So each evidence will be an exact string match of a sentence from the input document. We have also ensured while annotation that the annotators also selects the evidence from the document as shown in Figure \ref{fig:annotation}. This will help us measure the recall of encoder and classification agent against annotated data. The recall metric is defined as:

\begin{equation}
Recall = \frac{|t_{human} \cap t_{agent}|}{|t_{human}|}
\end{equation}

where $|t_{human} \cap t_{agent}|$ represents the number of tokens that intersect between the human annotator and the agent, and $|t_{human}|$ is the total number of tokens identified as evidence by the human annotator. This measures if the Jury Agent had enough information to conclude the judgement.

\begin{figure}[h]
    \centering
    \includegraphics[width=\columnwidth]{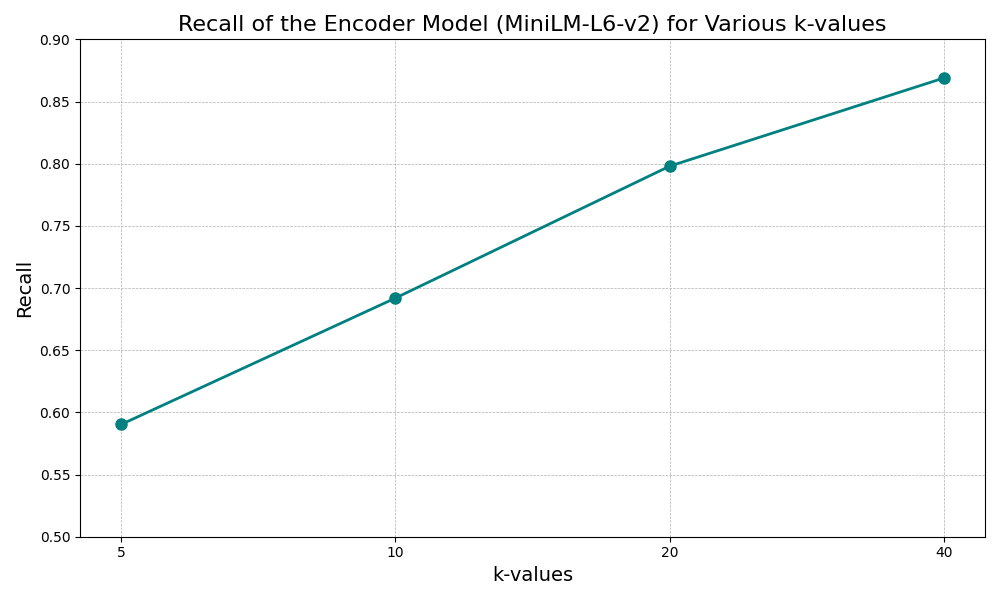}
    \caption{Recall of Encoder (MiniLM-L6-v2) model for various k-values}
    \label{fig:encoder}
\end{figure}

For encoder model, we used \texttt{MiniLM-L6} \footnote{Refer: https://huggingface.co/sentence-transformers/all-MiniLM-L6-v2}. We took the top similarity sentences given by encoder model for various k-values and calculated recall against the human evidences and computed the average recall for all checklist items. The results are plotted in Figure \ref{fig:encoder}. For $k = 40$, we get recall as 0.8689, which concludes that using encoder preserves useful information while discarding around 85\% of irrelevant data (average MIMIC data has 300 sentences) towards the judgement.

\begin{table}[H]
\centering
\caption{Recall metric for Classification Agent with different k values}
\label{tab:recall}
\begin{tabularx}{\columnwidth}{@{}lXX@{}}  
\toprule
\textbf{Model}         & \textbf{$\boldsymbol{k=10}$}     & \textbf{$\boldsymbol{k=20}$}    \\ \midrule
\textbf{GPT-4}         & 0.5792 & 0.6741 \\
\textbf{GPT-3.5}       & 0.4844 & 0.5554 \\
\textbf{Claude-Opus}   & 0.5254 & 0.5845 \\
\textbf{Calude-Sonnet} & 0.5042 & 0.5430  \\ \bottomrule
\end{tabularx}
\end{table}

On similar lines, we calculated the recall of the Classification Agent by comparing segregated evidences: if humans marked a checklist item as true, we compared the supporting evidences from the agent to those identified by humans, and similarly, if marked as false, we compared the contradictory evidences.  Table \ref{tab:recall} shows the recall of Classification Agent for various LLMs. Clearly for $k=20$ we have significantly higher recall as more evidences were present for the classifier to act upon. GPT-4 outperfomed other models with a maximum recall of 0.67 while other models showed slightly lower values.

\begin{figure}[h]
    \centering
    \includegraphics[width=\columnwidth]{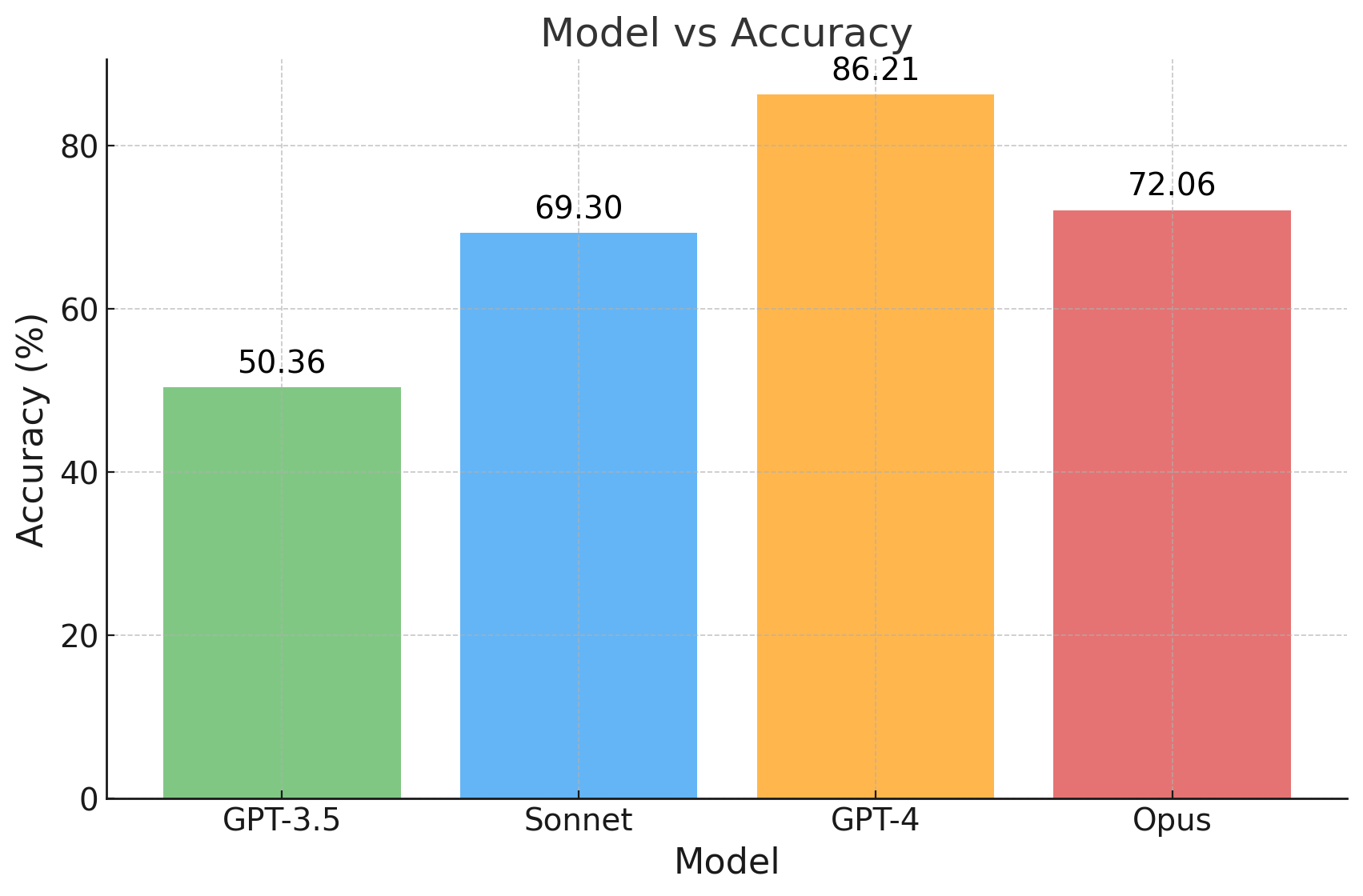}
    \caption{Accuracy of various LLMs for Jury Agent}
    \label{fig:accuracy_task1}
\end{figure}

\begin{figure}[h]
    \centering
    \includegraphics[width=\columnwidth]{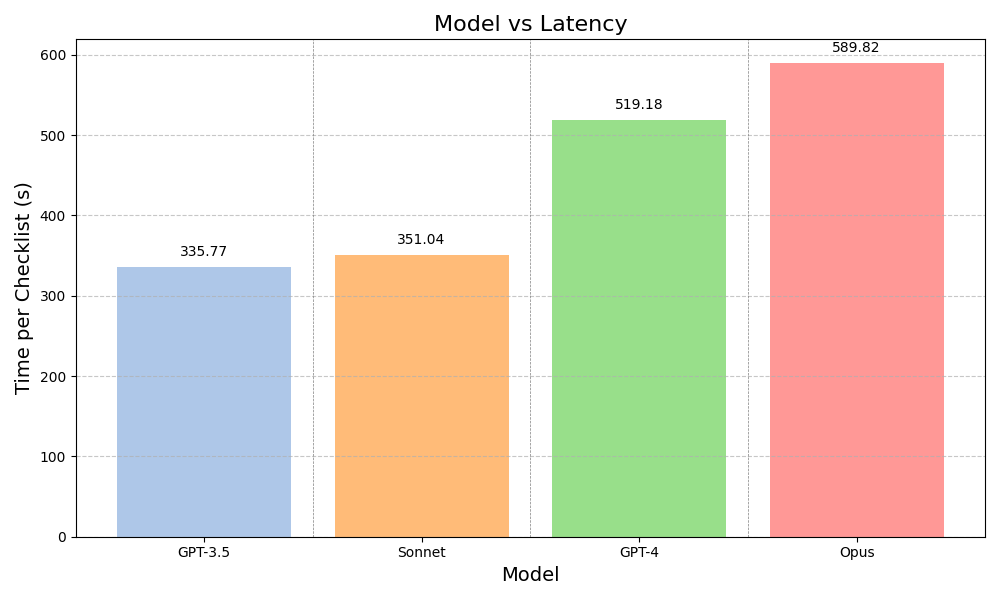}
    \caption{Latency of various LLM model for the leaf-node pipeline}
    \label{fig:latency_task1}
\end{figure}

We conducted a comprehensive evaluation of the Jury Agent ($\mathcal{M}_j$) employing various Large Language Models (LLMs), with a primary focus on accuracy and how it is affected with the number of retrieved evidences ($k$). GPT-4 and Opus demonstrated robust performance, achieving accuracies of 86\% and 72\% (Figure \ref{fig:accuracy_task1}), respectively. Notably, while Sonnet exhibited a slightly lower accuracy of 69\% compared to Opus, it provided a considerable advantage in terms of latency, reducing it by approximately 32\% (Figure \ref{fig:latency_task1}).

\begin{figure}[h]
    \centering
    \includegraphics[width=\columnwidth]{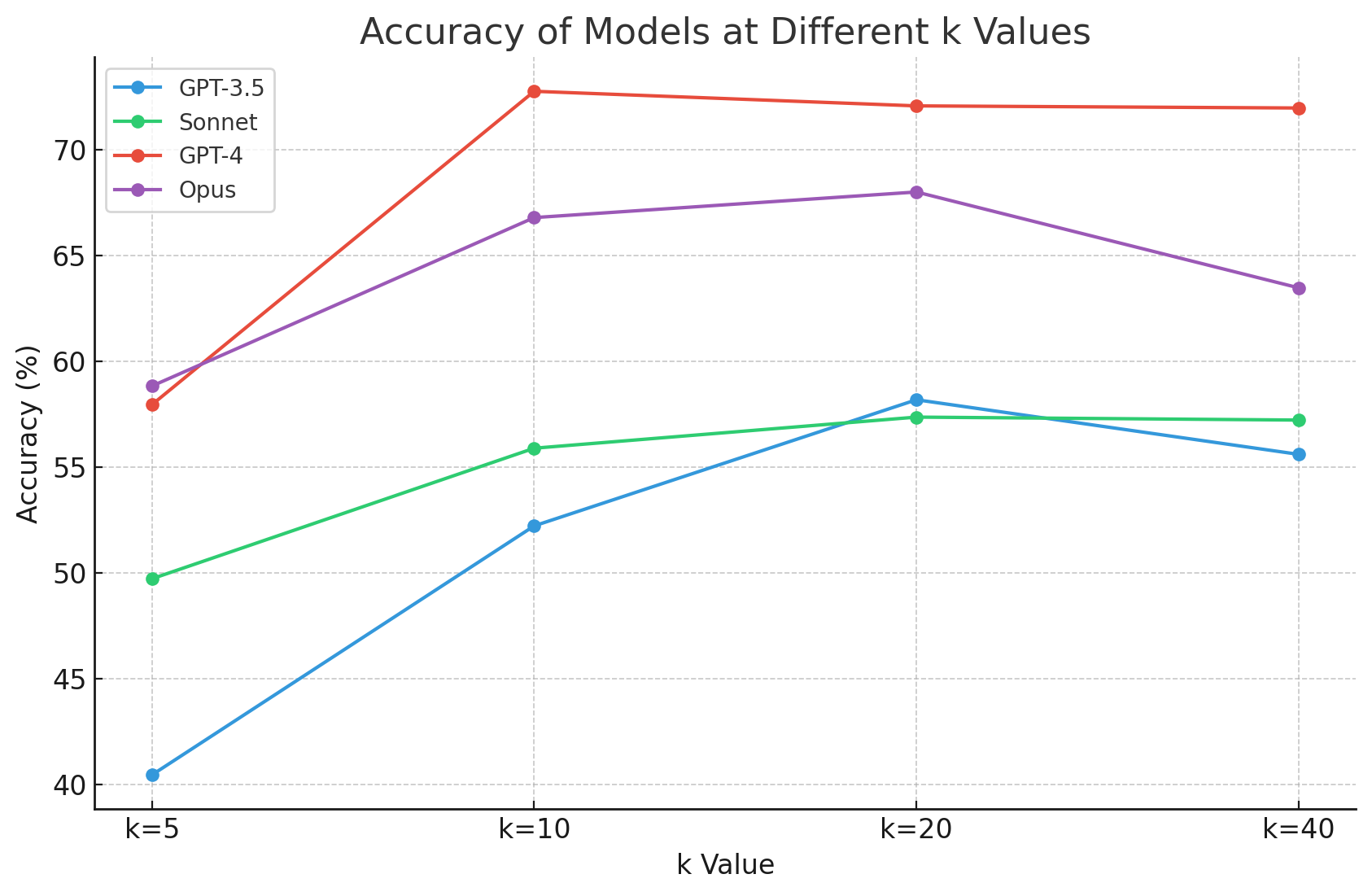}
    \caption{Effect of various k-values on Jury Agent}
    \label{fig:k_task1}
\end{figure}

\paragraph{Effect of Number of Retrieved Evidences (\textit{k}):} To better understand the effect of \textit{k} on our pipeline and choose the best value we tweaked the value of \textit{k} and ran the pipeline on a smaller sample of our dataset consisting of 20 checklists ( having 680 checklist items). We observed that as we increase the value of \textit{k}, the model performance increases till a value of $k=20$, after which the accuracy gets saturated as shown in Figure \ref{fig:k_task1}.

\begin{table*}
\centering
\caption{Model performance for Propagator Agent using Chain of Thought (CoT) \& In-Context Learning (ICL)}
\label{table:Comparison of model performance for Propagator Agent}
\resizebox{\textwidth}{!}{
\begin{tabular}{@{}lcccccccccc@{}}
\toprule
    & \multicolumn{2}{c}{\textbf{GPT-4}} & \multicolumn{2}{c}{\textbf{GPT-3.5}} & \multicolumn{2}{c}{\textbf{Claude-Sonnet}} & \multicolumn{2}{c}{\textbf{Claude-Opus}} \\
    \cmidrule(lr){2-3} \cmidrule(lr){4-5} \cmidrule(lr){6-7} \cmidrule(lr){8-9} \cmidrule(lr){10-11}
    \textbf{CoT + ICL} & \textbf{ICL} & \textbf{CoT + ICL} & \textbf{ICL} & \textbf{CoT + ICL} & \textbf{ICL} & \textbf{CoT + ICL} & \textbf{ICL} & \textbf{CoT + ICL} \\ 
\midrule
    \textbf{Response Accuracy (\%)} & 87.17 & 95.60 & 78.75 & 91.20 & 82.05 & 85.71 & 85.34 & 95.24 \\
    \textbf{Score Accuracy (\%)}    & 78.35 & 93.04 & 48.35 & 85.34 & 53.66 & 81.31 & 79.12 & 94.50 \\
    \textbf{Operator Accuracy (\%)}    & 89.27 & 95.01 & 81.04 & 92.82 & 82.05 & 87.54 & 84.78 & 94.04 \\
\bottomrule
\end{tabular}
}
\end{table*}

\subsection{Parent-node Judgement}

Our \textit{Propagator Agent} is an LLM-powered Agent, which takes up a parent node and its corresponding leaf nodes (along with their judgments and confidence scores) to obtain the judgment and confidence score of the parent node. This was done in two ways. In the first experiment, the LLM agent is asked directly to determine the response and score given parent statement and its child statements, responses and scores. The agent has to understand the logical operators (AND, OR, NOT) and then combine the child responses (True, False, No Information) to conclude parent judgement. The logical rules for No Information items is given in Figure \ref{fig:rules} and rules for calculating confidence score is given in Figure \ref{fig:scores}. In the second experiment, the LLM agent was asked to compute the logical operator between each child item and then the calculation of response and confidence score was done programmatically.

We evaluate the performance of the \textit{Propagator Agent} across various dimensions. The outcomes of this analysis are presented in Table \ref{table:Comparison of model performance for Propagator Agent}. The score accuracy refers to the accuracy of both the response and confidence score propagated correctly while the response accuracy is accuracy of only response being propagated correctly to the parent node resulting from the first experiment. The operator accuracy refers to the accuracy of the model to correctly identify the operators as done in the second experiment.

From the table we can conclude that the Agent is able to propagate response more accurately than confidence scores, as propagating confidence score is a more complex task than determining the response which involves only logical operations. Second experiment shows that the accuracy of operator determination task is comparable to the response accuracy determined using first approach. Once the operators are determined, response and confidence score are calculated programatically. Since determining operator would be a one time task (to be done while creating guidelines) taking second approach would get us similar accuracy but at significantly lower cost.

\begin{figure}[H]
\begin{tcolorbox}[
    arc=1mm, 
    colback=gray!10, 
    colframe=gray!80, 
    boxrule=0.5pt,
    fonttitle=\bfseries\scriptsize, 
    title=Rule Set for No Information Items,
    fontupper=\scriptsize\fontfamily{pcr}\selectfont, 
    left=3mm, 
    right=3mm, 
    top=2mm, 
    bottom=2mm 
]
\raggedright
\textbf{Case I: AND Operator}
\begin{enumerate}
  \item True \textbf{AND} No Information = No Information
  \item False \textbf{AND} No Information = False
\end{enumerate}

\textbf{Case II: OR Operator}
\begin{enumerate}
  \item True \textbf{OR} No Information = True
  \item False \textbf{OR} No Information = No Information
\end{enumerate}

\textbf{Case III: NOT Operator}
\begin{enumerate}
\item \textbf{NOT} No Information = No Information
\end{enumerate}

\end{tcolorbox}
\captionof{figure}{Rule Set for No Information Items followed by Propagator Agent for parent node judgement}
\label{fig:rules} 
\end{figure}

\paragraph{Effect of Prompting Strategy:} We performed two sets of experiments. The first involved providing \textit{In-Context Learning} (ICL) examples \cite{min2022rethinking} and measuring accuracy. Larger models such as GPT-4 and Opus yielded strong results, whereas smaller models like Sonnet and GPT-3.5 exhibited suboptimal performance when relying solely on ICL prompts. However, in the second experiment, when supplemented with \textit{Chain of Thought} (CoT) prompting \cite{NEURIPS2022_9d560961}, the performance of these smaller models markedly improved, demonstrating how the step-by-step reasoning process aids in decomposing the complex task of propagation into manageable segments. However, the use of Chain of Thought (CoT) prompting substantially increases response times for larger models due to its generation of an increased number of tokens compared to ICL-only prompting. In contrast, the enhancements in performance observed with GPT-3.5 are achieved without a marked increase in latency, particularly when compared to larger models such as Opus and GPT-4 under similar conditions.

\begin{figure}[h]
\begin{tcolorbox}[
    arc=1mm, 
    colback=gray!10, 
    colframe=gray!80, 
    boxrule=0.5pt,
    fonttitle=\bfseries\scriptsize, 
    title=Confidence Score ($f$) Calculation,
    fontupper=\scriptsize\fontfamily{pcr}\selectfont, 
    left=3mm, 
    right=3mm, 
    top=2mm, 
    bottom=2mm 
]
\raggedright
\textbf{Case I: AND Operator}
\begin{enumerate}[leftmargin=*]
  \item If final response is True:
  
  $f_{par}$ = min($f$ of all True child responses)
  \item If final response is False:
  
  $f_{par}$ = max($f$ of all False child responses)
  \item If final response is No Information:
  
  $f_{par}$ = min($f$ of all No Information child responses)
\end{enumerate}

\textbf{Case II: OR Operator}
\begin{enumerate}[leftmargin=*]
  \item If final response is True:
  
  $f_{par}$ = max($f$ of all True child responses)
  \item If final response is False:
  
  $f_{par}$ = min($f$ of all False child responses)
  \item If final response is No Information:
  
  $f_{par}$ = min($f$ of all No Information child responses)
\end{enumerate}

\end{tcolorbox}
\captionof{figure}{Confidence Score calculation rules followed by Propagator Agent for parent node judgement}
\label{fig:scores} 
\end{figure}

\paragraph{Effect of LLM Choice:} We conducted an evaluation of the \textit{Propagator Agent} utilizing various LLMs, with a particular emphasis on metrics such as accuracy and latency. Opus and GPT-4 emerged as the top performers, achieving approximately 94-95\% accuracy when CoT prompting was combined with ICL examples.

\begin{figure}[h]
    \centering
    \includegraphics[width=\columnwidth]{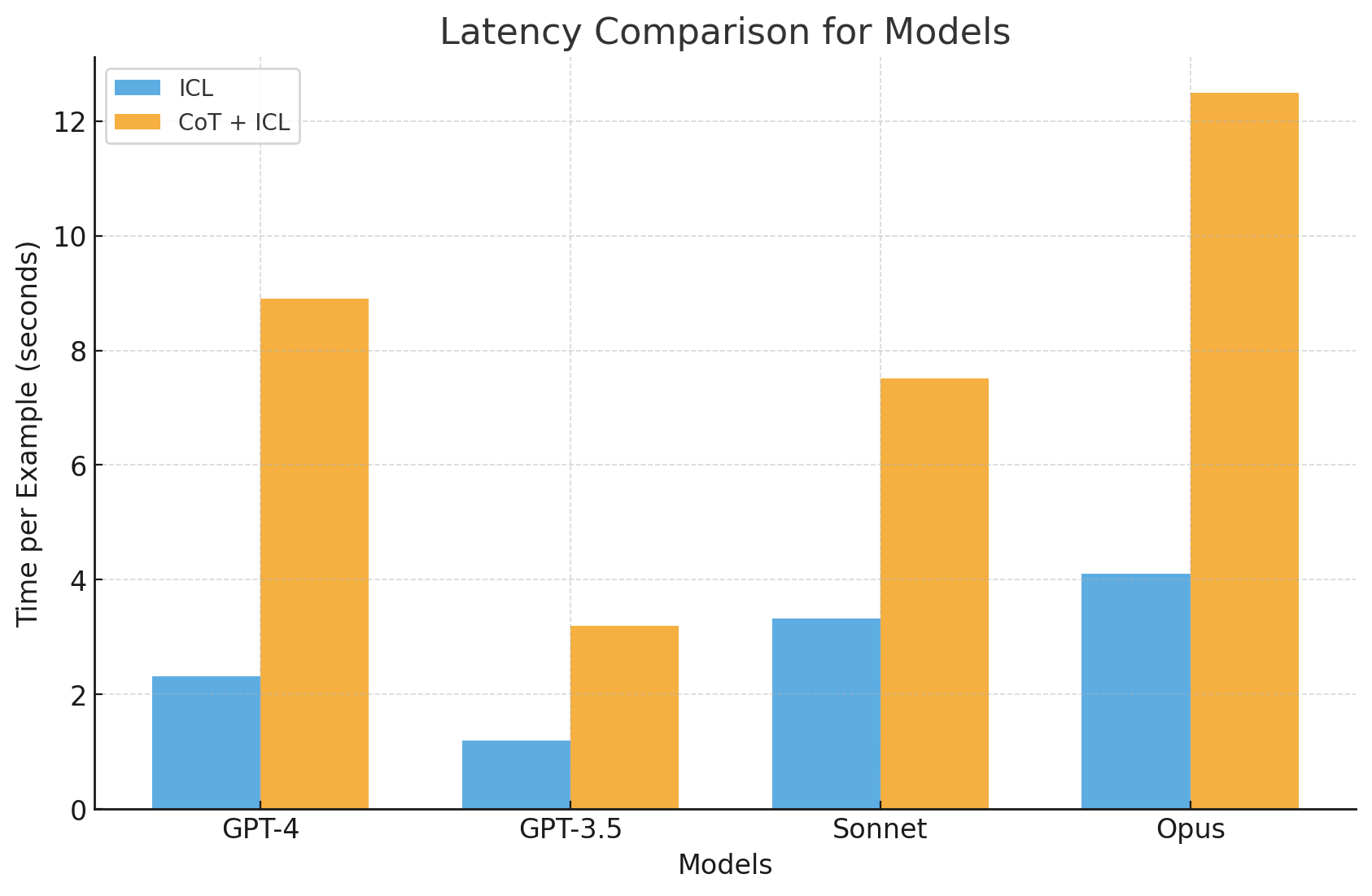}
    \caption{Latency Analysis of LLMs Under ICL and CoT for Propagator Agent when computing score accuracy}
    \label{fig:time}
\end{figure}

GPT-3.5 is ranked second in terms of accuracy but presents significant benefits in reduced latency compared to GPT-4 and Opus, as depicted in Figure \ref{fig:time}. Additionally, the operational costs associated with GPT-3.5 are substantially lower. Although selecting the optimal model involves a trade-off, GPT-3.5 stands out as the preferred option when considering a balance among cost, latency, and accuracy. Nonetheless, for scenarios where maximum accuracy is crucial, the larger models such as GPT-4 and Opus are more appropriate.

\section{Conclusion}

Our experiments utilized MIMIC-Note data, a set of string-based data. However, real-world applications typically involve obtaining resources (FHIR data) from EHR systems. Converting these resources into stringified data poses a unique engineering challenge. Although manageable, it is crucial to determine whether this data format could impact the effectiveness of our system.

In our approach, we integrated the use of confidence scores. Agents at the leaf nodes compute a confidence score for their predictions, which is then propagated up to the root node alongside the response. The confidence score at the root node is vital as it reflects the system’s certainty about the prediction quality. Checklists with low confidence scores are directed to a service layer where experienced professionals can review or adjust the model responses. This feedback loop can be leveraged to refine and enhance future models.

Given our focus on the healthcare sector, ensuring the explainability of outputs from these LLM agents was paramount. The decision-making process was elucidated through Chain of Thought (CoT) prompting and evidence collected by the Classification Agent, enhanced the transparency needed when AI models are employed in healthcare workflows.

While initially designed to automate prior authorization (PA) filing, this solution could also improve clinical decision support (CDS) systems by providing real-time alerts to physicians during consultations. For instance, it could alert physicians to incomplete medical records when prescribing treatments requiring PA, ensuring necessary documentation is promptly addressed. Thus, system responsiveness or latency becomes a critical metric for assessing its performance.

We have shown that breaking down a large, complex problem into smaller, specialized tasks handled by distinct agents can significantly enhance our ability to automate sophisticated tasks that were previously very challenging. This strategy also facilitates the shift from a monolithic AI solution ($\mathcal{M}$) to a micro-service architecture-driven solution ($\mathcal{M}_e$, $\mathcal{M}_j$, and $\mathcal{M}_p$). Currently, our method involves a constrained workflow, but it holds potential for evolving into a system with loosely coupled agents that are more dynamic and capable of improved problem-solving.

The ideal implementation of this methodology would adopt a structure akin to an organization, where the architecture consists of several pods. Each pod contains worker agents specialized in different aspects of the problem, complemented by checker agents that reassess and validate the outputs, triggering reruns when necessary. A super-orchestrator agent would oversee and coordinate the activities across the architecture. This setup aims to mitigate common issues like hallucination often seen in existing LLMs.

\section{Acknowledgments}
The authors wish to express their gratitude to Gaurav Grontayal and Tanmay Johri for their invaluable assistance in gathering the necessary checklists, hiring consultants and annotators, resolving annotation conflicts, and managing the overall data annotation pipeline. We also extend our thanks to RISA Labs for their support of our research. Special thanks are due to all annotators involved for their diligent work in data annotation.

\bibliography{acl_latex}

\end{document}